\documentclass[runningheads]{llncs}
\usepackage{graphicx}
\begin{document}
\title{Towards Simple and Efficient Task-Adaptive Pre-training for Text Classification}
\titlerunning{An Efficient Task-Adaptive Pre-training for Text Classification}
%
\author{Arnav Ladkat\inst{1,3}\thanks{Authors contributed equally} \and 
Aamir Miyajiwala\inst{1,3,*} \and 
Samiksha Jagadale\inst{1,3,*} \and 
Rekha Kulkarni\inst{1} \and 
Raviraj Joshi\inst{2,3}} 
\authorrunning{A. Ladkat et al.}
%
\institute{Pune Institute of Computer Technology, Pune, Maharashtra, India \and
Indian Institute of Technology Madras, Chennai, Tamilnadu, India
\email{ravirajoshi@gmail.com} \and
L3Cube, Pune, Maharashtra, India
}
\maketitle              
\begin{abstract} 
Language models are pre-trained using large corpora of generic data like book corpus, common crawl and Wikipedia, which is essential for the model to understand the linguistic characteristics of the language. New studies suggest using Domain Adaptive Pre-training (DAPT) and Task-Adaptive Pre-training (TAPT) as an intermediate step before the final finetuning task. This step helps cover the target domain vocabulary and improves the model performance on the downstream task. In this work, we study the impact of training only the embedding layer on the model's performance during TAPT and task-specific finetuning. Based on our study, we propose a simple approach to make the intermediate step of TAPT for BERT-based models more efficient by performing selective pre-training of BERT layers. We show that training only the BERT embedding layer during TAPT is sufficient to adapt to the vocabulary of the target domain and achieve comparable performance. Our approach is computationally efficient, with 78\% fewer parameters trained during TAPT. The proposed embedding layer finetuning approach can also be an efficient domain adaptation technique.

\end{abstract}

\section{Introduction}

\par Large-scale Pre-trained Language Models (PLMs) are extensively trained on massive heterogeneous datasets, known as pre-training datasets. These models are “Pre-trained”~\cite{devlin2018bert,howard2018universal,mccann2017learned,peters1802deep}, where they learn contextual representations by unsupervised learning methods like masked language modeling and next sentence prediction. Pre-training is followed by “Finetuning”, which uses supervised learning for tasks such as text-classification \cite{wagh2021comparative,khandve2022hierarchical}. The task for which finetuning is performed is called the downstream task. Previous works have shown that the representations learned from pre-training datasets help the model achieve a strong performance across multiple downstream tasks \cite{joshi2022l3cube,wani2021evaluating}.

\par
\begin{figure}[h]
    \centering
    \includegraphics[width=0.6\textwidth]{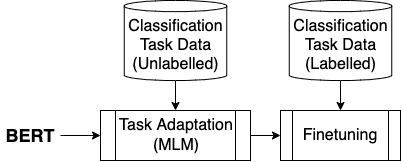}
    \caption{Representation of standard TAPT flow where pre-trained BERT is adapted to the target task using un-supervised MLM on task-specific data, followed by task-specific supervised finetuning.}
    \label{fig:tapt}
\end{figure}

 Contextual representations are typically generated from generic large-scale corpora-based pre-training, while downstream tasks are usually domain-specific. Recent work shows that additional pre-training performed on PLMs using in-domain and downstream task data improves the model's performance. Domain Adaptive Pre-training \cite{gururangan2020don,konlea2020domain} is a method used to achieve the same by continued pre-training of the PLM on a large corpus of unlabelled in-domain data. To expand on this approach, \cite{gururangan2020don,konlea2020domain} continued pre-training of language models on smaller unlabeled data drawn from the given task (Task-Adaptive Pre-training or TAPT) mirrors the gains associated with domain adaptation and can also be used as an additional pre-training step after DAPT to improve performance further. TAPT   (Fig.\ref{fig:tapt}) can also be used as an alternative to DAPT when sufficient in-domain data is unavailable in the worst-case scenario.

\par While the initial pre-training is essential for the model to understand and learn the linguistic characteristics of the English language through the encoder layer, DAPT and TAPT are essential for the model to get familiar with the domain-specific vocabulary without disturbing the domain-independent linguistic features learned by the encoder layer \cite{hewitt2019structural,jawahar2019does,liu2019linguistic}. With the motivation of adapting the pre-trained model to the target domain, we study the impact of training of only the embedding layer on the performance of the BERT-based model during TAPT. We specifically target the embedding layer as it encodes the token information. 

\par Although additional pre-training on in-domain or task data has shown to improve model performance on downstream tasks, it adds to the computational overhead and an indirect financial and environmental impact \cite{bannour-etal-2021-evaluating}. This work proposes a new method for efficient domain adaptation for BERT-based models. After the initial pre-training of the language model, during TAPT, we freeze the encoder layer and update only the embedding layer along with the task-specific dense layers. We show that this simple yet effective approach significantly reduces the number of trainable parameters and the time taken to train the model without impacting the model accuracy. The advantages of this restricted intermediate pre-training are twofold. It adapts the token embeddings to the target domain without forgetting the language characteristics learned by the upper BERT encoder layers during initial large-scale pre-training, thus preventing catastrophic forgetting  or overfitting \cite{howard2018universal} due to smaller task-specific datasets.

The main contributions of this work are as follows:
\begin{itemize}
\item This is the first work to evaluate embedding layer only fine-tuning during intermediate pre-training or language modeling stage. We show that this technique is an efficient training strategy to perform task adaptation during intermediate MLM pre-training.
\item During the task-specific fine-tuning stage keeping the embedding layer trainable yields superior performance as compared to freezing both BERT embedding and encoder layers. Although fine-tuning all the layers gives the best results, frozen encoder layers + trainable embedding layer may be desirable in low-resource settings.
\item Overall, we propose to only fine-tune the BERT embedding and task-specific layer, freezing the rest! This is constrained to the scenarios mentioned in the paper.
\end{itemize}

\section{Related Work}

\par Recently, Task-Adaptive Pre-training (TAPT) has become a popular topic for research, introduced by \cite{gururangan2020don}. It is essentially the adaptation of a Language Model (LM) to a target task leading to the improvement in model performance. Work done by \cite{li2021task} expands and discusses the effectiveness of TAPT  and its fusion with Self-training.

\par \cite{gururangan2020don} investigated the benefits of tailoring a pertained model like RoBERTa to the domain of a target task. Their work analyses four domains, namely biomedical and computer science publications, news and reviews, on eight classification tasks. This investigation is further extended into the transferability of adapted language models across all the tasks and domains. Finally, a study of the significance of pre-training on human-curated data is carried out.

\par The study proposed in \cite{konlea2020domain} discusses various strategies to adapt BERT and DistilBERT to historical domains and tasks exemplary for computational humanities. The results encourage the integration of continued pertaining into machine learning tasks for improved performance stability. A combination of domain adaptation and task adaptation shows positive effects. When applied individually, task adaptation can be performed in every setup, unlike domain adaptation, where sufficient in-domain data is necessary.

Several approaches have been undertaken to make TAPT more efficient, especially with methods involving word embeddings.

A study in \cite{nishida2021task} focuses on the static word embeddings of the pre-trained language models for domain adaptation. The researchers propose a  process called Task-Adaptive Pre-training with word Embedding Regularization (TAPTER) to teach pre-trained language models the domain-specific meanings of words. Word embeddings in the target domain are obtained by training a fastText model on the downstream task training data. These word embeddings are made close to the static word embeddings during TAPT. TAPTER performs better than the standard finetuning and TAPT when in-domain data is deficient in the initial pre-training dataset.

Another method is proposed by \cite{el2021domain} for specializing in general-domain embeddings in a low-resource context. More specifically, they have considered a worst-case scenario where only the target task corpus is available. Given the availability of general-domain models which can be used to initialize training for any specialized domain, it is concluded that re-training from a general model is less expensive and leads to comparable, although slightly lower performance.

Researchers propose an alternative approach in \cite{sachidananda2021efficient} for transferring pre-trained language models to new domains by adapting their tokenizers. It is shown that adaptive tokenization on a pre-trained RoBERTa model provides more than 97\% of the performance benefits of domain-specific pre-training. However, this approach incurs a 6\% increase in model parameters due to the introduction of 10,000 new domain-specific tokens.

Methods to make TAPT and DAPT more efficient, as discussed above, involve methods such as using static pre-trained in-domain embeddings, adapting the tokenizer and training fast-text or word2vec models on in-domain data. These approaches introduce multiple extra steps to the pre-training of the PLMs, whereas our approach leverages BERT's static embeddings matrix. Given a specialized target domain, we aim to improve the quality of general-domain word representations using in-domain corpora.

\section{Experimentation Setup}

This section discusses the experimental setup followed to study the effect of restricting training to the embedding layer during TAPT and finetuning. This setup is also used to evaluate the proposed efficient domain or task adaptation method.

\subsection{Datasets}

We have used four benchmark text classification datasets. Firstly, the \textbf{IMDB}\footnote{https://huggingface.co/datasets/imdb} dataset comprises 50K highly polar movie reviews, 25K for training and 25K for testing, with positive and negative classification labels. Second is \textbf{AG-News}\footnote{https://huggingface.co/datasets/ag\_news}, a topic classification dataset containing news articles on four classes: World, Sports, Business, and Science. Each class includes 30K training samples and 1,900 testing samples, with a total of 120K training samples and 7,600 testing samples. The third is the \textbf{Emotion}\footnote{https://huggingface.co/datasets/emotion} dataset, made of English Twitter messages with six primary emotions: anger, fear, joy, love, sadness, and surprise. It contains 16K training data samples and 2K samples for validation and testing each. Lastly, we have \textbf{BBC News}\footnote{http://mlg.ucd.ie/datasets/bbc.html}, the smallest dataset in our experiment, comprising News Articles across five domains. It consists of 16K training samples, 450 testing samples, and 150 samples for validation.

\subsection{Model}

In the experiments, we use Bidirectional Encoder Representations from Transformers (BERT) model \cite{devlin2018bert} for MLM and finetune it for text classification on the target dataset. The BERT model consists of 12 layers of bidirectional transformer-based encoder blocks, where each layer has 12 self-attention heads. BERT base uncased\footnote{https://huggingface.co/bert-base-uncased} is pre-trained on a large English corpus \cite{wolf2020transformers} in a self-supervised fashion with two objectives - Masked Language Modeling (MLM) and Next Sentence Prediction (NSP). 

\subsection{Methodology}

\begin{figure}[h]
    \centering
    \includegraphics[width=0.7\textwidth]{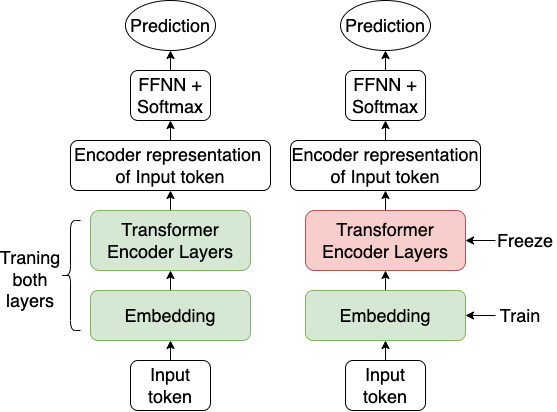}
    \caption{The left model depicts the standard TAPT flow, whereas right model indicates the proposed TAPT approach where BERT encoder layers are frozen during intermediate pre-training.}
    \label{fig:model}
\end{figure}

The standard approach to performing TAPT for BERT-based models involves training all layers on training data of the downstream task. Our approach involves freezing the encoder layer while training only the embedding and final task-specific dense layers (Fig.\ref{fig:model}).
By doing so, we specialise the general domain word representations according to the target tasks.

\begin{table}
\centering
\caption{Results of different configurations during restricted finetuning.}\label{tab:tapt-table}
\begin{tabular}{|l|l|l|l|l|}
\hline
\textbf{TAPT} & \textbf{IMDB} & \textbf{AG-News} & \textbf{Emotion} & \textbf{BBC-News} \\
\hline
 None  & 92.6 & 89.67 & 93.4 & 96.4 \\
 Standard  & \textbf{93.19} & \textbf{89.76} & 92.85 & \textbf{97.08} \\
 Freeze Encoder Layer & 93.02 & \textbf{89.78} & \textbf{93.3} & \textbf{97.08} \\
\hline
\end{tabular}
\end{table}

\begin{table}
\centering
\caption{Results of different configurations during restricted finetuning.}\label{tab:finetuning-results}
\begin{tabular}{|l|l|l|l|l|l|}
\hline
 \textbf{TAPT} & \textbf{Finetuning} & \textbf{IMDB} & \textbf{AgNews} & \textbf{Emotion} & \textbf{BBC-News} \\
\hline
None & Standard & 92.6 & 89.67 & 93.4 & 96.4 \\
 None & Freeze Encoder Layer & 89.06 & 88.96 & 86.85 & 93.93 \\
 None & Freeze Encoder and Embedding  & 83.82 & 85.27 & 56.85 & 91.23 \\
 Standard & Standard & 93.19 & 89.76 & 92.85 & 97.08 \\
 Standard & Freeze Encoder Layer & 91.02 & 90.44 & 88.3 & 95.05 \\
 Standard & Freeze Encoder and Embedding  & 86.42 & 83.25 & 54.1 & 93.25 \\
\hline
\end{tabular}
\end{table}

\subsection{Evaluation Setup}

\subsubsection{Restricted TAPT}
To evaluate our proposed approach, we defined three experimental setups.
\begin{itemize}
    \item The first setup is the baseline, where we perform standard finetuning on the pre-trained BERT model using a target classification task, where no task adaptation is performed.
    \item Secondly, we evaluate the effectiveness of the standard task-adaptive pre-training using the unlabelled data of the target classification dataset. We train all the layers of the model, followed by standard finetuning.
    \item Lastly, the model is pre-trained using our approach for Task Adaptation, updating only the embedding layer and freezing the entire encoder block, followed by standard finetuning.
\end{itemize}

The results are shown in the Table \ref{tab:tapt-table}.

\subsubsection{Restricted Finetuning} Our experiments also explore the impact of selective training of layers during the model's final finetuning. We explore three modes of finetuning - standard full finetuning, freezing the encoder plus embedding layer, and freezing the encoder. Each of these experiments is preceded with and without TAPT giving a total of six configurations. The results and description are mentioned in Table \ref{tab:finetuning-results}.

\section{Results}

As observed in Table \ref{tab:tapt-table}, after using the proposed approach for TAPT, the model's accuracy is comparable to the standard approach. Moreover, restricted TAPT shows a slight improvement in the performance for three of the four datasets. There is a 78\% drop in trainable parameters using our approach, resulting in significant improvement in the time taken per epoch during TAPT. The difference in time taken for TAPT can be seen in Table \ref{tab:time-results}.

We also study the impact of training the embedding layer during final finetuning. The results of the six configurations are shown in Table \ref{tab:finetuning-results}. We observe that finetuning the embedding layer gives better results than its frozen counterpart. Finetuning all the layers gives the best performance, followed by finetuning embedding and dense layer (frozen encoder setup), further followed by finetuning only dense layer (frozen encoder and embedding layer setup).

\begin{table}
\centering
\caption{Training time taken per epoch in minutes. The timings were computed on Intel(R) Xeon(R) CPU @ 2.20GHz and Tesla P100 GPU.}\label{tab:time-results}
\begin{tabular}{|l|l|l|}
\hline
\textbf{Dataset} & \textbf{Standard TAPT} & \textbf{Proposed Approach}\\
\hline
IMDB & 56.7 & 38.0 \\
 Ag-News & 37.63 & 9.23 \\
 Emotion & 1.03 & 0.5 \\
 BBC-News & 2.05 & 1.43\\
\hline
\end{tabular}
\end{table}

\subsection{Conclusion}
In this work, we study the impact of training only the embedding layer of the BERT model during task adaptation and finetuning and, based on the findings, propose an approach to perform TAPT efficiently. We adapt the token embeddings of the transformer model to the target task by updating only the embedding layer and freezing the encoder layers, thus retaining the domain-independent linguistic features. The approach is evaluated on four benchmark text classification datasets. We observe that the model performance is not negatively impacted, and we prevent overfitting and catastrophic forgetting on small datasets by only training  21\% of the model parameters during task adaptation. The approach also significantly reduces the training time per epoch.

\section*{Acknowledgements}
This work was done under the L3Cube Pune mentorship
program. We would like to express our gratitude towards
our mentors at L3Cube for their continuous support and
encouragement.

%

\bibliographystyle{splncs04}
\bibliography{main}

\end{document}